# Robust Behavioral Cloning for Autonomous Vehicles using End-to-End Imitation Learning

Tanmay Vilas Samak, Chinmay Vilas Samak and Sivanathan Kandhasamy

***Abstract –*** *In this work, we present a lightweight pipeline for robust behavioral cloning of a human driver using end-to-end imitation learning. The proposed pipeline was employed to train and deploy three distinct driving behavior models onto a simulated vehicle. The training phase comprised of data collection, balancing, augmentation, preprocessing and training a neural network, following which, the trained model was deployed onto the ego vehicle to predict steering commands based on the feed from an onboard camera. A novel coupled control law was formulated to generate longitudinal control commands on-the-go based on the predicted steering angle and other parameters such as actual speed of the ego vehicle and the prescribed constraints for speed and steering. We analyzed computational efficiency of the pipeline and evaluated robustness of the trained models through exhaustive experimentation during the deployment phase. We also compared our approach against state-of-the-art implementation in order to comment on its validity.*

## 1. INTRODUCTION

Autonomous driving [1] is one of the exponential technologies of the current century and has been a dream of mankind since a long time. There are fundamentally two approaches to tackle this problem viz. the old school robotics-based approach [2] and the modern learning-based approach [3].

The traditional robotics-based approach splits the autonomous driving task into subsequent modules, namely perception, planning and control. Although this somewhat simplifies the overall process, precise implementation of these individual fragments is an arduous task in itself. Furthermore, the complex algorithms pertaining to perception (particularly machine vision), planning (specifically online planning) and control (especially optimal controllers) are computationally expensive and often struggle from real-time processing; not to mention they are generally scenario-specific and need to be retuned before being deployed to handle dissimilar situations.

With the advent of machine learning strategies, some of the aspects of robotics approach were substituted using alternative data driven algorithms. Particularly, convolutional neural networks (CNNs) [4] completely revolutionized the way perception stack was implemented. Nonetheless, the notion of end-to-end learning [5] truly turned the tables by defining the entire task of autonomous driving as a machine learning problem. End-to-end learning, in the context of autonomous driving, allows an agent to directly map the perception data to the appropriate actuator commands using neural network as a non-linear function approximator. This eliminates the need of designing and implementing specialized perception, planning and control stacks, which not only simplifies the development phase but also gives an added advantage of real-time computation during the deployment phase.

This research focuses on end-to-end imitation learning aimed at autonomous driving, and although this technology has numerous advantages as discussed earlier, it has its own limitations as well. Apart from general limitations such as long training times, requirement of labelled datasets, patience of tuning hyperparameters and inability of surpassing trainer's performance, there are some significant pitfalls to this technology. First of all, the trained models cannot generalize beyond a certain limit, and the fact that their safety/reliability cannot be guaranteed restricts this approach at the research level for now. Secondly, training a model for end-to-end longitudinal control using solely camera frames is extremely difficult due to its dependence on several other parameters. Finally, this approach hasn't been much demonstrated in complex driving scenarios, such as high-density traffic or intersections. Nevertheless, autonomous driving in high-density traffic is very well achievable using a similar approach since the traffic/pedestrians may be treated as dynamic obstacles and a neural network may be made to learn to avoid colliding with them; similar to [6]. For the task of intersection navigation, a behavioral planner may be trained in an end-to-end manner to turn the vehicle in the appropriate direction. Possible changes may include additional sensing modalities or adoption of a hybrid autonomous driving software stack.

The task of cloning driving behavior of a human being using the end-to-end imitation learning approach has been accomplished by experts in the field. Pomerleau [7] was one of the earliest to demonstrate end-to-end learning for lateral motion control of an autonomous vehicle; however, owing to the technological constraints back in 1989,

T.V. Samak, C.V. Samak and S. Kandhasamy are with the Autonomous Systems Lab, Department of Mechatronics Engineering, SRM Institute of Science and Technology, Kattankulathur 603203, Tamil Nadu, India. Email: {tv4813, cv4703, sivanatk}@srmist.edu.in

the work adopted a fully connected neural network, which is really small according to present standards. Muller, et. al. [8] successfully applied end-to-end learning to train a 6-layer CNN to teleoperate a scaled radio controlled (RC) vehicle through an off-road obstacle course. Building on top of [8], Bojarski, et.al. [9] trained a 9-layer CNN to map the raw pixels from a single camera frame directly to the steering angle. In [10], Bojarski, et.al. described the salient features learnt by the 9-layer CNN described in [9] and explained how the network predicted steering angles in an end-to-end manner. Xu, et. al. [11] trained a long-short term memory fully convolutional network (LSTM-FCN) using a large-scale crowd-sourced dataset. Given present camera frame and past egomotion states, the trained model could predict a distribution over future egomotion, which was a significant accomplishment in terms of multi-agent behavioral cognizance.

It is to be noted that most of the existing implementations make use of extensively large datasets to clone human driving behaviors. Additionally, they make use of adequately deep neural network architectures in order to impart generalization capability to the model. Both of these significantly increase the training time and question the efficiency of the pipeline in terms of training robust models within a short time.

It is a common observation that the training performance (i.e. training speed and/or accuracy) is predominantly affected by data collection methods [12], followed by the choice of neural network architecture and hyperparameter values. Additionally, the number of steps employed in preprocessing the data before feeding it to the neural network affects the training time and deployment latency, equally. The goal is to, therefore, design a pipeline that is not only temporally efficient in terms of training and deployment, but is also able to impart sufficient robustness to the models being trained.

This research lays the foundational work towards developing a lightweight pipeline for robust behavioral cloning, which bridges the gap between training performance and robustness of a driving behavior model trained using end-to-end imitation learning. Particularly, this work proposes one such pipeline, which is aimed at faster and efficient training while also imbibing the necessary robustness to the model against environmental variations. The pipeline also ensures a low deployment latency with the focus of real-time implementation. We adopt the said pipeline to clone three distinct driving behaviors and analyze its performance through a set of experiments specifically aimed at testing the robustness of the trained models. We also compare the performance of our pipeline against NVIDIA's state-of-the-art implementation [9] in order to comment on its validity.

## 2. BACKGROUND

This section describes the contextual details pertaining to implementation and analysis of the proposed pipeline.

### 2.1. Simulation System

The simulation system employed for validating the proposed pipeline was a modified version of an open-source simulator developed by Udacity [13]. The modifications included altering the environment (changing some of the existing objects, adding new objects, varying lighting conditions, etc.) as well as the vehicle (replacing vehicle body, tuning vehicle dynamics, varying position, orientation, camera count, etc.) along with the implementation of data logging functionality and an organized graphical user interface (GUI).

The simulator was developed atop the Unity [14] game engine so as to simulate accurate system dynamics (which predominantly affected vehicle motion control) and detailed graphics (which mimicked acquisition of realistic perception data through simulated cameras). It is therefore possible to implement the proposed approach explicitly on hardware or as a sim2real application.

### 2.2. Driving Scenarios

This work describes cloning of three specific driving behaviors in order of increasing complexity, namely simplistic driving, rigorous driving and collision avoidance. Each behavior model was trained and deployed in a dedicated simulated environment designed specifically for the respective behavior (Figure 1). It is to be noted that the sequence of training these behaviors does not matter since all the behaviors were trained afresh, without any prior knowledge or experience.

The simplistic driving scenario was aimed at training the ego vehicle to drive around a race track, with a rather smooth profile, a few sharp turns and a bridge with drastically different road texture. The vehicle was to remain in the drivable portion of the road and drive smoothly for the entire length of the track.

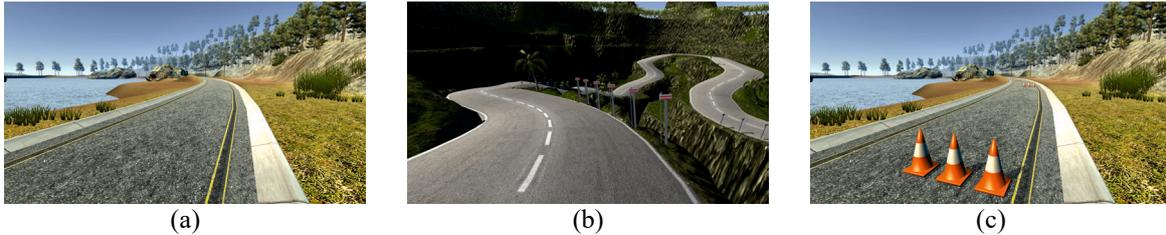

*Figure 1. Simulated driving scenarios for training and testing (a) simplistic driving behavior, (b) rigorous driving behavior and (c) collision avoidance behavior*

Taking this a level further, the ego vehicle was trained in the rigorous driving scenario to specifically learn maneuvering across a very curvy and bumpy mountainous track, with a few blind turns included. Additionally, the scenario contained significantly high number of shadows, which is reported to drastically affect the perception system [15]. The vehicle was to also observe lane-keeping while traversing the track autonomously.

The collision avoidance scenario was designed as an additional challenge, wherein the ego vehicle was made to revisit the race track, only this time it had to traverse through numerous static obstacles placed randomly throughout the course. Standard traffic cones were used as obstacles and were placed such that approximately half the road was blocked at a time. In order to stretch the comfort zone, the vehicle was trained with just a single camera. Additionally, the number and position of obstacles was varied during the deployment phase to test the generalization capability of the trained model, and hence the robustness of the pipeline. The vehicle was to remain in the drivable portion of the road and avoid collision with any of the obstacles by maneuvering away from them. Furthermore, the vehicle was to drive in a rather smooth fashion when no obstacles were present on the course.

## 2.3. Experiments

A set of experiments was framed in order to test the robustness of the driving models trained using the proposed pipeline. The degree of autonomy $\eta$ exhibited by the ego vehicle was computed based on the ratio of interference time $t_{int}$ and the total lap time $t_{lap}$ (Eq. 1).

$$\eta \; (\%) = \left[ 1 - \frac{t_{int}}{t_{lap}} \right] * 100 \tag{1}$$

Each interference was assumed to take up a total of 6 seconds [9], implying a direct relation between the interference time $t_{int}$ and the total number of interferences $n_{int}$ during a complete lap (Eq. 2).

$$t_{int} = n_{int} * 6 \tag{2}$$

Following is a list of said experiments along with their respective objectives:

1.  ***No Variation:*** The deployment scenario was kept identical to the training scenario so as to validate the autonomy of the vehicle without any scenic variations. Results were reported in terms of degree of autonomy exhibited by the ego vehicle.
2.  ***Scene Obstacle Variation:*** The number of static obstacles in the scene was varied between 20 (during training), 10 and 0. Additionally, the position and orientation of the obstacles was also changed in each case. Results were reported in terms of degree of autonomy exhibited by the ego vehicle. This experiment was carried out only for the collision avoidance behavior.
3.  ***Scene Light Intensity Variation:*** The intensity of scene light was varied with increments of $\pm 0.1$ cd w.r.t. the original value. This experiment tested robustness of the trained model against variation in brightness of camera frame. Results were reported in terms of upper and lower limits of variation for which the ego vehicle exhibited ~100% autonomy.
4.  ***Scene Light Direction Variation:*** The direction of scene light was varied about the local X-axis w.r.t. the original value with an angular resolution of $\pm 1°$. This experiment tested robustness of the trained model against variation in shadows. Results were reported in terms of upper and lower limits of variation for which the ego vehicle exhibited ~100% autonomy.
5.  ***Vehicle Position Variation:*** The spawn location of vehicle was set to a different position as compared to that during data collection. This experiment tested robustness of the trained model against variation in initial conditions. Results were reported in terms of degree of autonomy exhibited by the ego vehicle.

6. ***Vehicle Orientation Variation:*** The orientation of the vehicle was varied with increments of $\pm5°$ about the local Y-axis[1] w.r.t. its original value. This experiment tested robustness of the trained model in terms of converging back to the lane center. Results were reported in terms of upper and lower limits of variation for which the ego vehicle exhibited ~100% autonomy.

7. ***Vehicle Heading Inversion:*** The vehicle was spawned facing opposite direction of the track, implying an orientation shift of 180° about the local Y-axis. This experiment tested the generalization capability of the trained model. Results were reported in terms of degree of autonomy exhibited by the ego vehicle.

8. ***Vehicle Speed Limit Variation:*** The speed limit of the vehicle was increased with increments of 5 km/h w.r.t. the original value of 30 km/h during data collection. This experiment critically tested the deployment latency as the model was required to predict steering angles at a faster rate. Results were reported in terms of upper limit of variation for which the ego vehicle exhibited ~100% autonomy.

# 3. IMPLEMENTATION

The implementation of proposed pipeline can be divided into two phases, viz. training phase and deployment phase (Figure 2). The following sections discuss each phase along with the specifics for each driving scenario.

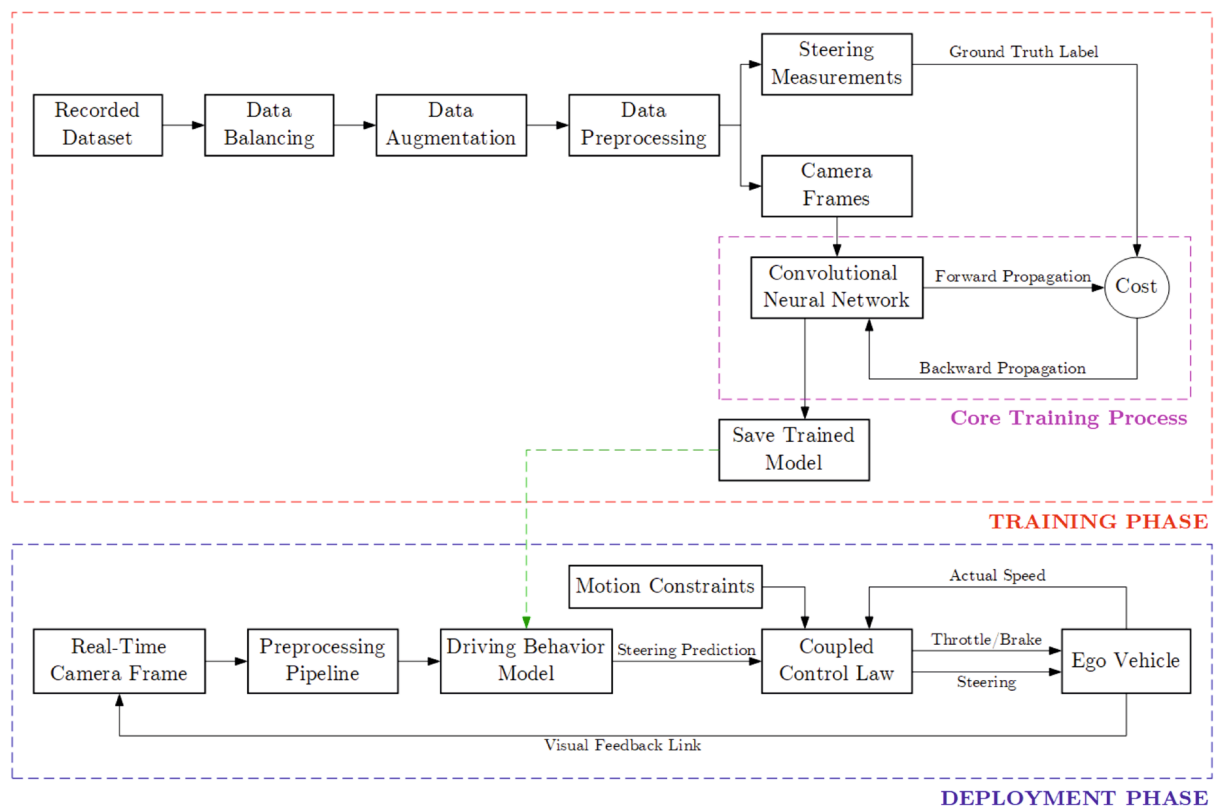

*Figure 2. Architecture of the proposed pipeline distinctly illustrating the training and deployment phases*

## 3.1 Training Phase

Training phase comprised of data collection, balancing, augmentation, preprocessing and training neural network.

### 3.1.1. Data Collection

Independent datasets were collected by manually driving the ego vehicle (using a standard computer keyboard) across the three driving scenarios described earlier. The manual driving task was accomplished by first and second authors so as to reduce biased driving behavior. While data collection for training simplistic driving behavior included 10 laps of manual driving, that for training rigorous driving and collision avoidance behaviors included 20 laps of manual driving, each.

---

[1] Unity employs a left-handed co-ordinate system with X-axis pointing right, Y-axis pointing up and Z-axis pointing forward.

The datasets for simplistic and rigorous driving behaviors included timestamped frames from the center, left and right cameras onboard the vehicle and the normalized steering angle measurement corresponding to each timestamp. On the other hand, the dataset for collision avoidance behavior included timestamped frames from the center camera alone and the normalized steering angle measurement corresponding to each timestamp.

*Table 1. Data collection details*

| Parameter | Specification |
|---|---|
| Simulator resolution | 1920×1080 px (Full HD) |
| Targeted simulator frame rate | 60 FPS |
| Simulated camera resolution | 320×160 px |
| Simulated camera field of view | 60° |
| Data collection rate[2] | 1.5 Hz |

The common details pertaining to dataset collection are summarized in Table 1. It is to be noted that in addition to camera frames and steering angle measurements, the dataset also contained normalized measurements of throttle and brake commands as well as the vehicle speed corresponding to each timestamp; however, these measurements were only used for comparative analysis of simulated field results (refer section 4.4).

### 3.1.2. Data Segregation

The collected datasets were randomly split into training and validation subsets in the ratio of 4:1 (i.e. 80% training data and 20% validation data). Table 2 holds the number of data samples contained within the said datasets.

*Table 2. Data segregation details*

| Driving Behavior | Data Samples | | |
|---|---|---|---|
| | *Complete Dataset* | *Training Dataset* | *Validation Dataset* |
| Simplistic Driving | 12101 | 9680 | 2421 |
| Rigorous Driving | 50911 | 40728 | 10183 |
| Collision Avoidance | 25471 | 20376 | 5095 |

The random state of splitting each dataset was chosen specifically such that the training and validation datasets would have minimal variation w.r.t. the steering measurements.

### 3.1.3. Data Balancing

The original training datasets were skewed towards either left or right steering since the ego vehicle traversed the track in a single direction. Additionally, all the collected datasets were heavily unbalanced towards zero-steering owing to the fact that the steering angle was reset to zero whenever the control keys were released. In order to minimize these unbalances, the dataset was balanced by adopting the following techniques.

#### 3.1.3.1. Skewed-Steering Unbalance Correction

The skewed-steering unbalance was balanced by employing the following strategies. In case of a severe unbalance, the ego vehicle was manually driven in the opposite direction of the track, thus compensating for unequal number of turns in either direction. As an additional compensation, a flip augmentation technique was employed (refer section 3.1.4.4). Table 3 describes the skewed-steering unbalance correction(s) applied for the three driving behaviors.

*Table 3. Skewed-steering unbalance correction*

| Driving Behavior | Skewed-Steering Unbalance Correction |
|---|---|
| Simplistic Driving | Bi-directional driving (5 laps in one direction, 5 in other) + data augmentation (flip) |
| Rigorous Driving | None[3] |
| Collision Avoidance | Data augmentation (flip) |

---

[2] Although higher data collection rates were achievable, temporally proximal camera frames would be very similar and thus, wouldn't provide any novel information to the network.
[3] The dataset for rigorous driving behavior was negligibly skewed and hence bi-directional driving was not carried out. Furthermore, data augmentation (flip) technique couldn't be applied to this scenario since flipping of the frame also changed the lane of the ego vehicle (which conflicted with lane-keeping objective).

### *3.1.3.2. Zero-Steering Unbalance Correction*

In order to deal with the zero-steering unbalance, a random portion of the dataset containing exactly zero steering angle measurements was deleted at each pass. It is to be noted that while a high prejudice towards zero-steering may affect the generalization capability of trained model, a significant one is still required in order to impart smooth driving ability to the model. The amount of data to be deleted $D$ was defined relative to the total number of zero-steering measurements $d$ in the entire dataset (Eq. 3).

$$D = \lfloor d * \lambda \rfloor \qquad (3)$$

Note that the deletion rate $\lambda \in [0,1]$ is a hyperparameter, which was tuned independently for each driving behavior by analyzing the steering histogram and regulating the ratio of zero-steer to extreme-steer values based on the amount of aggressiveness required for that behavior. Table 4 describes the zero-steering unbalance correction applied for the three driving behaviors.

*Table 4. Zero-steering unbalance correction*

| Driving Behavior | Zero-Steering Unbalance Correction | |
| --- | --- | --- |
| | *Deletion Rate* | *Steering Histogram* |
| Simplistic Driving | 0.7 |  |
| Rigorous Driving | 0.8 |  |
| Collision Avoidance | 0.8 |  |

### 3.1.4. Data Augmentation

Supervised learning ideally mandates the training data to cover all the possible action-value pairs within the operational design domain (ODD) of the system being trained. However, collecting such an ideal dataset isn't always feasible. Following this notion, data augmentation was adopted to ensure robust training and correct any inherent unbalances within the datasets (refer section 3.1.3). It is to be noted that data augmentation was carried out offline during the core training phase and not while data collection. The simulated environment conditions were static throughout the data collection step.

In this work, a total of six augmentation techniques, viz. perspective shifts, shadows, brightness, flip, pan and tilt (in that exact sequence[4]) were applied to the dataset during the training phase. The probability of applying any particular augmentation was determined by a random variable $X \sim U(0,1)$ such that $p(0 < X \le x) = x$. Table 5 holds the probabilities of a specific augmentation being applied to a given data sample for each of the three driving behaviors.

---

[4] The sequence of applying augmentations is extremely important. Some augmentations may conflict with others, thereby producing absurd results.

*Table 5. Augmentation probabilities*

| Augmentation Technique | Probability of Application | | |
|---|---|---|---|
| | *Simplistic Driving* | *Rigorous Driving* | *Collision Avoidance* |
| Perspective Shifts | 0.50 | 0.50 | 0.00 |
| Shadows | 0.30 | 0.30 | 0.30 |
| Brightness | 0.40 | 0.40 | 0.40 |
| Flip | 0.50 | 0.00 | 0.50 |
| Pan | 0.10 | 0.10 | 0.10 |
| Tilt | 0.05 | 0.05 | 0.05 |

The following sections thoroughly explain each of the discussed augmentation techniques.

### 3.1.4.1. Perspective Shifts

Perspective shifts simulate high cross-track error, collecting actual data of which shall practically require unethical or potentially unsafe manual driving. This is accomplished by feeding the neural network with side camera frames (as if they were center camera frames) and correcting the corresponding steering angle labels to account for the synthetic cross-track error (Figure 3).

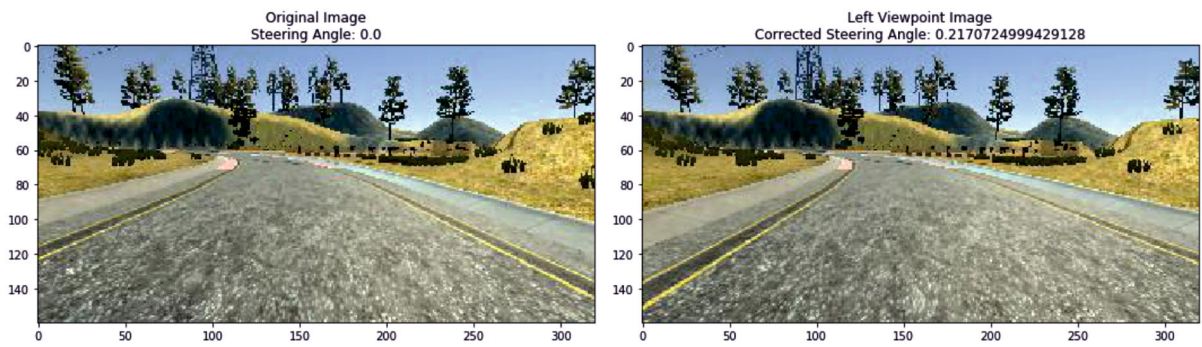

*Figure 3. Perspective shift augmentation applied to a sample camera frame from simplistic driving dataset*

The probability of applying perspective shift was defined to be 0.5 and the left and right viewpoints were further assigned equal selection probabilities, i.e. 0.25 each.

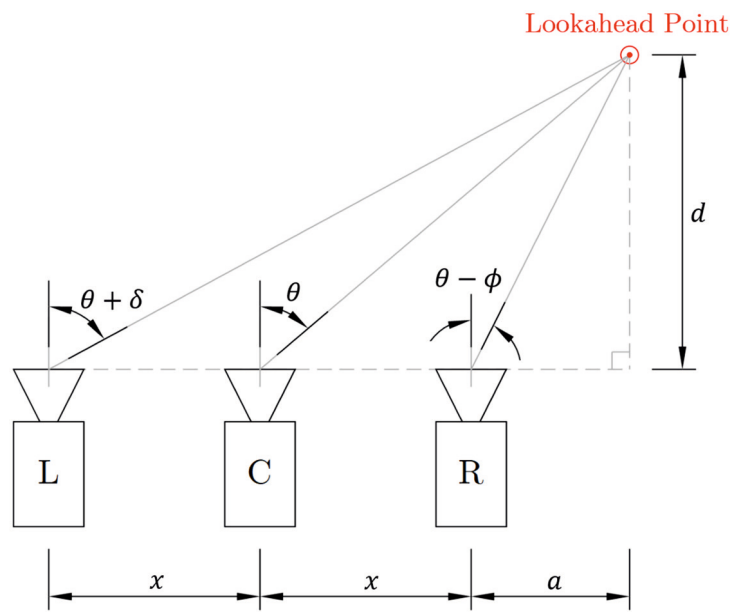

*Figure 4. Geometry of a 3-camera perspective shift augmentation system*

The correction angles $\delta$ and $\phi$ for left and right perspective shifts respectively are formulated as follows:

$$\delta = tan^{-1}\left[\frac{\gamma}{1 + tan^2(\theta) + \gamma * \tan(\theta)}\right] \qquad (4)$$

$$\phi = tan^{-1}\left[\frac{\gamma}{1 + tan^2(\theta) - \gamma * \tan(\theta)}\right] \qquad (5)$$

Note that $\gamma$ in Eq. 4 and 5 is the ratio of inter-camera distance $x$ and recovery distance $d$ as depicted in Figure 4. The implementations discussed in this work considered a constant recovery distance of 10 m and vehicle width of 1.9 m. The side cameras were assumed to be mounted on the left and right extremities of the vehicle body, thereby implying an inter-camera distance of 0.95 m.

### 3.1.4.2. Shadows

Synthetic shadows were generated (Figure 5) with an aim of imparting shadow-immunity to the trained model. Four quadrangular shadows with darkness coefficient (0.65) matching the shade of actual scene shadows were added to the camera frames. The vertices of all the polygons were chosen from discrete uniform distributions within the specified region of interest (lower half of the frame), i.e. $X{\sim}U(0, 320)$ and $Y{\sim}U(80, 160)$.

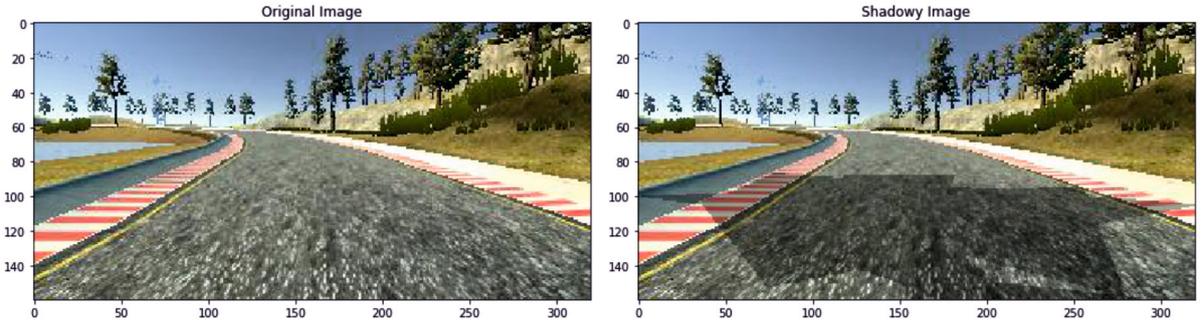

*Figure 5. Synthetic shadow augmentation applied to a sample camera frame from simplistic driving dataset*

### 3.1.4.3. Brightness

Image brightness drastically affects the feature extraction process since over and under exposure of a camera frame may deteriorate some of the critical features. Hence, it is strongly recommended to include examples of varied lighting conditions (Figure 6) within the training dataset, especially when cameras are the only sensing modality.

Each pixel of the image was added with a constant bias $\beta$ sampled from a uniform distribution $U(-100, 100)$; where negative values indicate darkening and positive indicate brightening. Finally, the pixel values were clamped in the range of $[0, 255]$.

$$dst_{i,j} = \max\left[0, \min\left[(src_{i,j} + \beta), 255\right]\right] \qquad (6)$$

The terms $src$ and $dst$ in Eq. 6 denote source and destination images respectively and the subscripts $i$ and $j$ indicate the location (row, column) of a specific pixel within the two images.

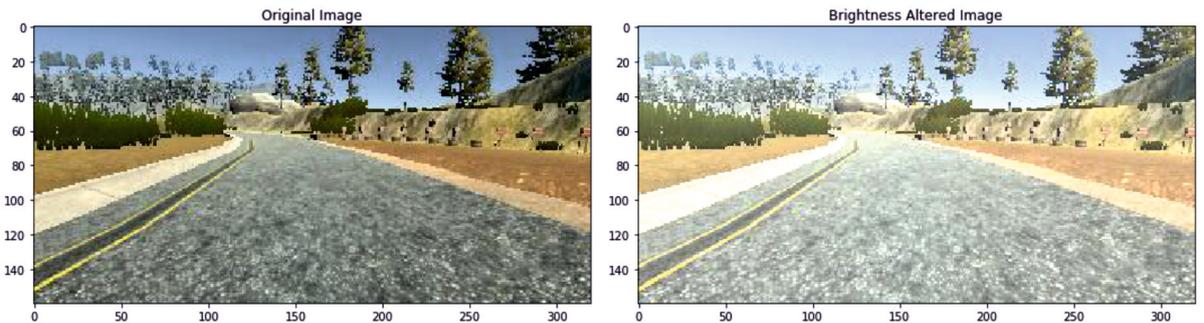

*Figure 6. Variable brightness augmentation applied to a sample camera frame from simplistic driving dataset*

### 3.1.4.4. Flip

As described in section 3.1.3.1, a flip augmentation technique was employed for reducing the effect of unbalanced steering angle distribution of a particular training dataset. This technique involved horizontal flipping of center camera frames (Figure 7) and negating the corresponding steering angles to compensate for the flip. Each frame had an equal chance to be flipped (i.e. $p_{flip} = 0.5$), thereby producing nearly equal number of opposite turns in the augmented dataset. It is to be noted that this augmentation was not applied to the side camera frames as left and right viewpoints were interchanged after flipping, leading to erroneous steering correction.

$$dst_{i,j} = src_{w-i-1,j} \tag{7}$$

$$\theta_t = -\theta_t \tag{8}$$

The terms $src$ and $dst$ in Eq. 7 denote source and destination images respectively of size $(w, h)$ and the subscripts $i$ and $j$ indicate the location (row, column) of a specific pixel within the two images. The variable $\theta_t$ in Eq. 8 depicts steering angle $\theta$ at discrete time instant $t$.

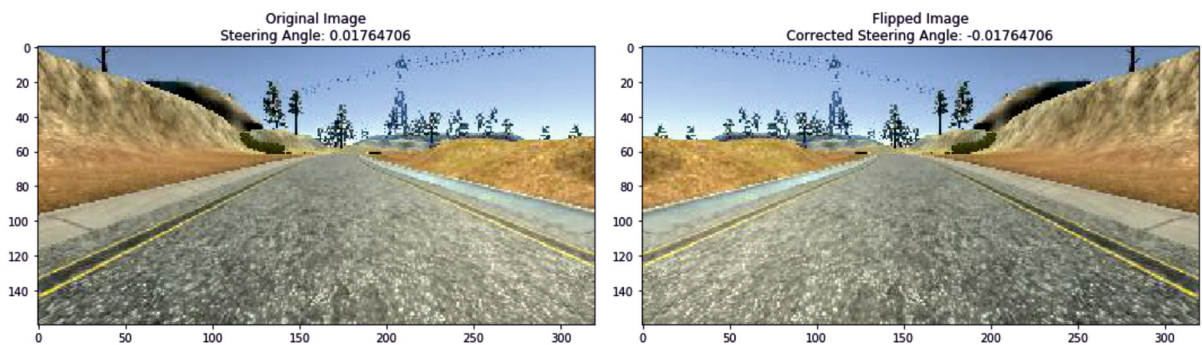

*Figure 7. Horizontal flip augmentation applied to a sample camera frame from simplistic driving dataset*

### 3.1.4.5. Pan

The panning operation (Figure 8) comprised of shifting the image pixels horizontally and/or vertically through a random amount relative to the original image dimensions, the magnitude of which was sampled from a uniform distribution $U(-0.05, 0.05)$. The transformation matrix $M_T$ for translating an image by $t_x$ and $t_y$ respectively in x and y directions is defined in Eq. 9.

$$M_T = \begin{bmatrix} 1 & 0 & t_x \\ 0 & 1 & t_y \end{bmatrix} \tag{9}$$

The null area resulting from panning was cropped out and the resulting image was resized to original dimensions.

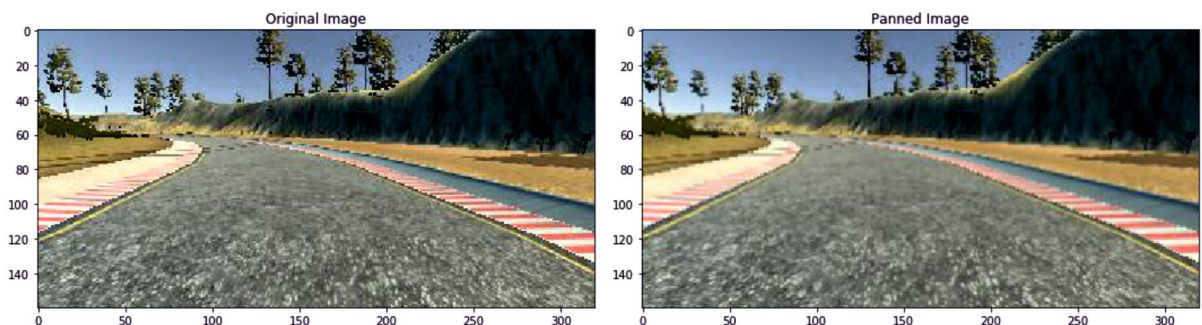

*Figure 8. Panning augmentation applied to a sample camera frame from simplistic driving dataset*

### 3.1.4.6. Tilt

The tilting operation (Figure 9) comprised of rotating the image about its center by a random angle (in degrees) sampled from a uniform distribution $U(-1, 1)$. The transformation matrix $M_R$ for rotating an image of size $(w, h)$ by an angle $\varphi$ about its center is defined in Eq. 10.

$$M_R = \begin{bmatrix} \cos(\varphi) & \sin(\varphi) & \frac{w}{2} * [1 - \cos(\varphi)] - \frac{h}{2} * \sin(\varphi) \\ -\sin(\varphi) & \cos(\varphi) & \frac{w}{2} * \sin(\varphi) + \frac{h}{2} * [1 - \cos(\varphi)] \end{bmatrix} \qquad (10)$$

The null area resulting from tilting was removed by cropping out the largest (maximal area) axis-aligned central rectangular region of interest (ROI) and resizing it back to original image dimensions. The dimensions $(w_{roi}, h_{roi})$ of the said ROI were computed based on the original image dimensions $(w, h)$ and the tilt angle $(\varphi)$ using the following relation:

$$w_{roi}, h_{roi} = \begin{cases} \dfrac{h}{2 * \sin(\varphi)}, \dfrac{h}{2 * \cos(\varphi)}; & \text{Half-constrained case} \\ \dfrac{w * \cos(\varphi) - h * \sin(\varphi)}{\cos(2\varphi)}, \dfrac{h * \cos(\varphi) - w * \sin(\varphi)}{\cos(2\varphi)}; & \text{Fully-constrained case} \end{cases} \qquad (11)$$

The half-constrained case in Eq. 11 implies two crop corners touching the longer side of the rotated image and the other two crop corners on a line joining midpoints of the shorter sides of the rotated image. Conversely, the fully-constrained case indicates all four crop corners touching the sides of rotated image.

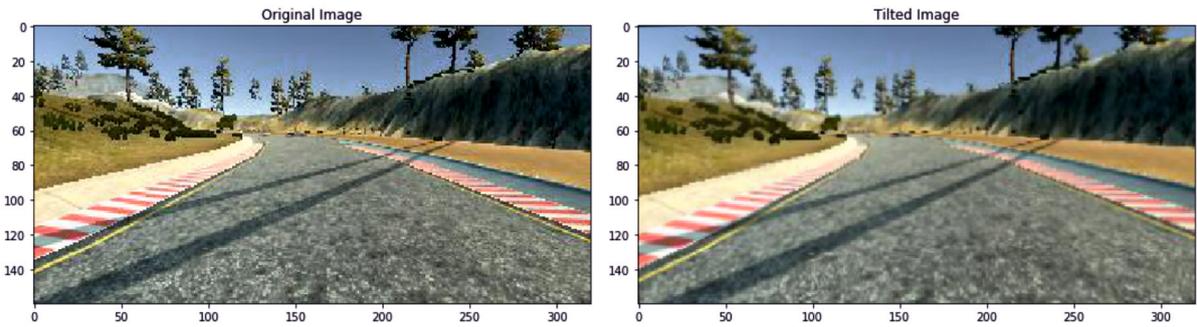

*Figure 9. Tilt augmentation applied to a sample camera frame from simplistic driving dataset*

### 3.1.5. Data Preprocessing

Data preprocessing was aimed at faster and efficient training as well as deployment. This work describes a two-step preprocessing function, which performs resizing and normalization (with mean centering) operations on the input images (Figure 10).

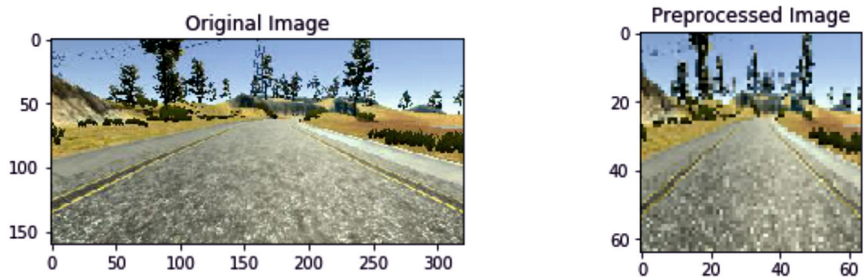

*Figure 10. Preprocessing steps applied to a sample camera frame from simplistic driving dataset*

### 3.1.5.1. Resizing

Resizing operation scaled down the 320×160 px input image by a factor of $f_x = 0.2$ and $f_y = 0.4$ respectively to yield a 64×64 px image, thereby also altering the aspect ratio. While downscaling enhanced the training rate, obtaining a 1:1 aspect ratio permitted effective usage of square kernels[5].

---

[5] Steering angle prediction relies predominantly on the horizontal information within an image, which follows that, for a non-square image, kernels must be rectangular too. Instead, we resize the image to 1:1 aspect ratio, thereby squeezing the horizontal information, and adopt square kernels.

### 3.1.5.2. Normalization and Mean-Centering

Normalization generally leads to faster convergence by speeding up the learning process. The resized frames $I: \{\mathbb{X} \subseteq \mathbb{R}^3\} \mapsto \{0, \cdots, 255\}$ with intensity values in range $[0, 255]$ were normalized to $I_N: \{\mathbb{X} \subseteq \mathbb{R}^3\} \mapsto \{0, \cdots, 1\}$ with intensity values in range $[0, 1]$ using the following relation (Eq. 12).

$$I_{N_{i,j}} = \frac{I_{i,j}}{255} \tag{12}$$

Additionally, the normalized images were mean-centered to zero by subtracting 0.5 from each pixel (Eq. 13).

$$I_{N_{i,j}} = I_{N_{i,j}} - 0.5 \tag{13}$$

Note that the subscripts $i$ and $j$ in Eq. 12 and 13 indicate the location (row, column) of a specific pixel within the respective images.

### 3.1.6. Training

While the data collection and segregation operations were performed only once, others such as data balancing, augmentation and preprocessing were executed on-the-go during training phase. It also shuffled and balanced the data samples after each pass through the training dataset, thereby ensuring that almost all the collected samples were fed to the neural network. Additionally, it preprocessed the validation data samples during validation phase.

### 3.1.6.1 Neural Network Architecture

The proposed training pipeline can be flexibly adopted for any neural network architecture. We adopt a relatively shallow CNN with 14065 trainable parameters to test the efficiency of the proposed pipeline, since smaller neural networks are reported to inherently possess limited generalization capability. The network was designed to accept 64×64 px RGB image as input and predict the required steering angle in an end-to-end manner.

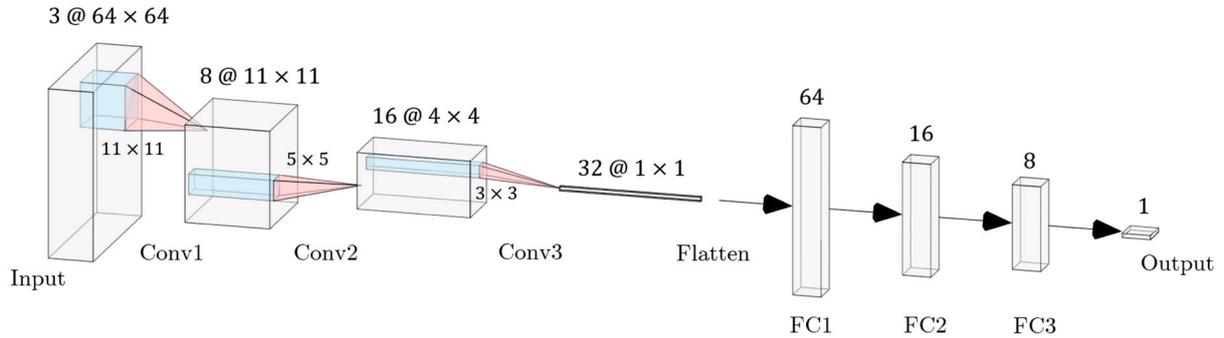

*Figure 11. Neural network architecture*

The network architecture consisted of 3 convolutional (Conv) and 3 fully connected (FC) layers (Figure 11). The convolutional layers performed strided convolutions with a 5×5 stride in the first layer and a 2×2 stride in the following two layers. The kernel size for first layer was set large[6], 11×11, and was progressively reduced thereafter with 5×5 in the second layer and 3×3 in the third. The output of third convolutional layer was flattened and passed through the 3 fully connected layers, each followed by a dropout [17], to ultimately predict the steering angle.

It is to be noted that the longitudinal controller was implemented independently to generate the throttle/brake commands on-the-go based on the predicted steering angle and other parameters such as actual speed of the ego vehicle and the prescribed constraints for speed and steering (refer section 3.2.2).

### 3.1.6.2 Hyperparameters

Training hyperparameters across the three driving scenarios were kept similar, with the only exception of number of epochs and steps per epoch. Table 6 summarizes the hyperparameters chosen for each driving scenario.

---

[6] Large kernels in the initial layers are reported to better extract the environmental parameters [16].

*Table 6. Training hyperparameters*

| Hyperparameter | Value | | |
|---|---|---|---|
| | *Simplistic Driving* | *Rigorous Driving* | *Collision Avoidance* |
| Activation Function | ReLU[7] | ReLU | ReLU |
| Dropout probability | [0.25, 0.25, 0.25] | [0.25, 0.25, 0.25] | [0.25, 0.25, 0.25] |
| Weight initialization | Glorot Uniform [18] | Glorot Uniform | Glorot Uniform |
| Bias initialization | Zeros | Zeros | Zeros |
| Optimizer | Adam [19] | Adam | Adam |
| Loss | MSE[8] | MSE | MSE |
| Learning rate | 1E-3 | 1E-3 | 1E-3 |
| Epochs | 5 | 10 | 5 |
| Batch size | 256 | 256 | 256 |
| Augmentation loops[9] | 64 | 64 | 64 |
| Training steps per epoch[10] | 2420 | 10182 | 5094 |
| Validation steps per epoch[11] | 10 | 40 | 20 |

## 3.2 Deployment Phase

Deployment phase comprised of data preprocessing and motion control.

### 3.2.1. Data Preprocessing

Deployment phase utilized the exact same two-step preprocessing pipeline as described in section 3.1.5. The operations included resizing the live camera frames from 320×160 px to 64×64 px and then normalizing and mean-centering them. The number of preprocessing operations were limited with an aim of minimizing the deployment latency in order to ensure real-time execution of the autonomous control loop (section 4.1 furnishes the necessary computational details).

### 3.2.2. Motion Control

The trained neural network model predicted instantaneous lateral control command $\theta$ (i.e. steering angle). On the other hand, a novel coupled control law (Eq. 14) was defined for generating the longitudinal control command $\xi$ (i.e. throttle and brake) based on the predicted steering angle $\theta$, actual vehicle speed $v_a$ and the prescribed speed and steering limits $v_l$ and $\delta$, respectively. Table 7 summarizes the boundary conditions for this novel coupled control law.

$$\xi = \tau \left[ \frac{(v_l - v_a)}{v_l} - \frac{|\theta|}{\delta} \right] \tag{14}$$

Note that $\tau$ in Eq. 14 is a proportionality constant, which controls the aggressiveness of longitudinal command. It can take values in range $[0, 1]$ (this work assumed $\tau = 1$ for all the described experiments).

*Table 7. Boundary condition analysis for coupled longitudinal control law*

| Boundary Condition | $|\theta| = 0$ | $|\theta| = \delta$ |
|---|---|---|
| $v_a = 0$ | $\xi = \tau$ | $\xi = 0$ |
| $v_a = v_l$ | $\xi = 0$ | $\xi = -\tau$ |

It is to be noted that positive and negative values of $\xi$ influence the throttle and brake commands respectively.

---

[7] Rectified Linear Unit (ReLU) is a non-linear activation function defined as $f(x) = \max(0, x)$, where $x$ is the input to a particular neuron.

[8] Mean Squared Error (MSE) loss over $n$ training examples is defined as $L(\hat{y}, y) = \frac{1}{n} \sum_{i=1}^{n} \|y_i - \hat{y}_i\|^2$, where $\hat{y}_i$ is the predicted output and $y_i$ is the corresponding ground truth label.

[9] Augmentation loops $\chi$ is an indicative of the number of times training data samples were fed to the neural network, with random augmentations applied each time.

[10] Training steps per epoch parameter was defined as $\left\lceil \frac{n}{m} \right\rceil * \chi$, where $n$ is number of training samples, $m$ is batch size and $\chi$ is number of augmentation loops.

[11] Validation steps per epoch parameter was defined as $\left\lceil \frac{n}{m} \right\rceil$, where $n$ is number of validation samples and $m$ is batch size.

# 4. RESULTS

## 4.1. Computational Details

The proposed pipeline was designed and implemented on a personal computer incorporating Intel i7-8750H CPU and NVIDIA RTX 2070 GPU, running Python 3.6.8 with TensorFlow-GPU 1.14.0. Table 8 holds the computational details pertaining to training as well as deployment phases of the pipeline. The training time w.r.t. the number of training epochs and MSE loss corresponding to each driving behavior is reported in second column. The third column, on the other hand, reports the latency of one cyclic execution of the entire pipeline during the deployment phase, starting from image preprocessing to steering and corresponding throttle command generation.

*Table 8. Computational details of the proposed pipeline*

| Driving Behavior | Training Time (hr) [# Data Samples, # Epochs, MSE Loss] | Deployment Latency (ms)[12] |
|---|---|---|
| Simplistic driving | ~1.4 [12101, 5, 1.7E-3] | |
| Rigorous driving | ~10.9 [50911, 10, 6.43E-2] | 1.5 to 3.0 |
| Collision avoidance | ~2.9 [25471, 5, 2.75E-2] | |

## 4.2. Activation Visualization

Prior to any other form of validation, the activations of each convolutional layer of the trained model were visualized to confirm whether the network had actually learned to detect significant environmental features, solely based on the steering angle labels. It was observed with utmost care that the feature map was activated in response to some significant environmental attributes such as road boundaries, lane markings or obstacles in the environment, depending upon the scenario.

The following figure illustrates, for each driving scenario, a sample preprocessed image fed to the neural network (left) followed by activation maps of the first, second and third convolutional layer. It is to be noted that the sample image was randomly selected from the training dataset.

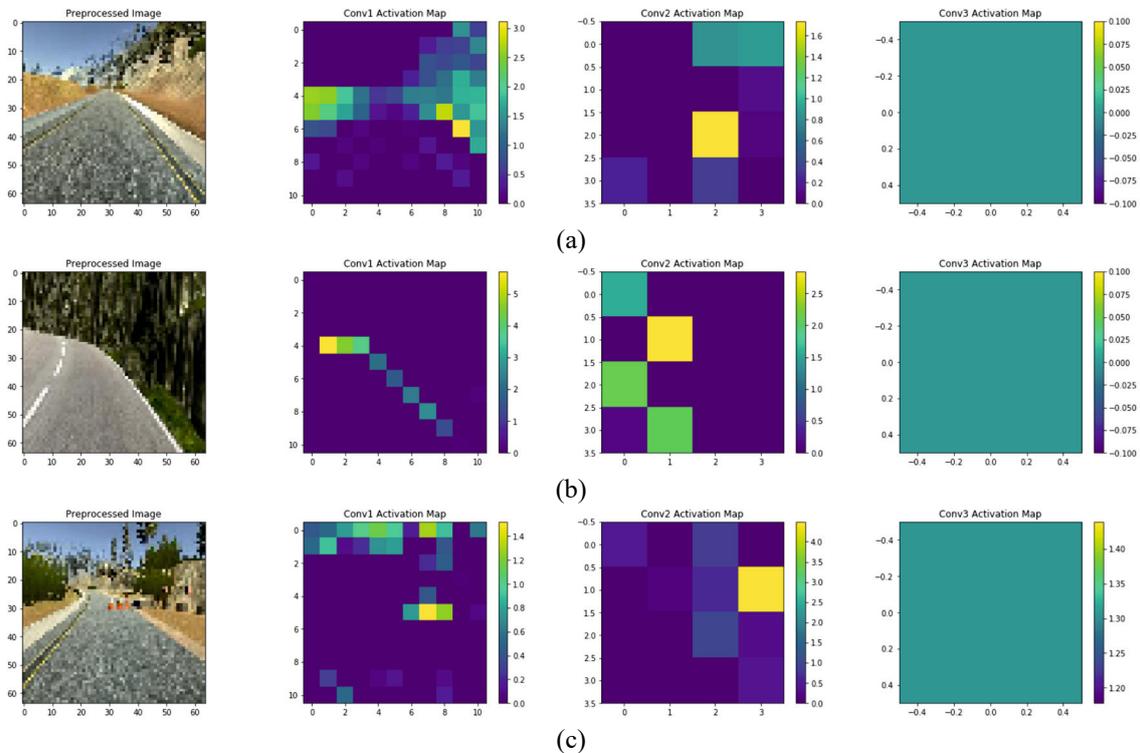

(a)

(b)

(c)

*Figure 12. Activation maps of CNN models trained for (a) simplistic driving, (b) rigorous driving and (c) collision avoidance behaviors*

---

[12] The statistical mode value for deployment latency was ~2 milliseconds.

It was observed that simplistic driving behavior model predominantly detected road boundaries and generated moderate activations for lane markings (Figure 12 (a)). Rigorous driving behavior model, on the other hand, clearly detected solid lane markings and produced minor activations for dashed lane markings separating the two driving lanes (Figure 12 (b)). Finally, the collision avoidance behavior model exhibited very powerful activations for obstacles and mild ones for lane markings (Figure 12 (c)). A general observation was that, irrespective of the driving scenario, high activations were an indicative of the non-drivable areas within the camera frame, or the boundary separating the drivable and restricted areas.

## 4.3. Prediction Analysis

After analyzing the activation maps and confirming that the neural network had indeed learned to detect significant environmental features from preprocessed camera frames, its ability to make intelligent steering predictions based on the learned features was validated through direct comparison against the ground truth labels within the manual driving dataset.

For this purpose, the models were fed with subset of the training data containing camera frames for approximately 1 lap and the steering angle predictions corresponding to each input camera frame were recorded. These recorded predictions were then plotted against manual steering commands corresponding to the respective frames obtained during data recording (Figure 13).

It was a significant observation that the trained model could produce highly smooth steering transitions and was able to track the general profile of ground truth labels.

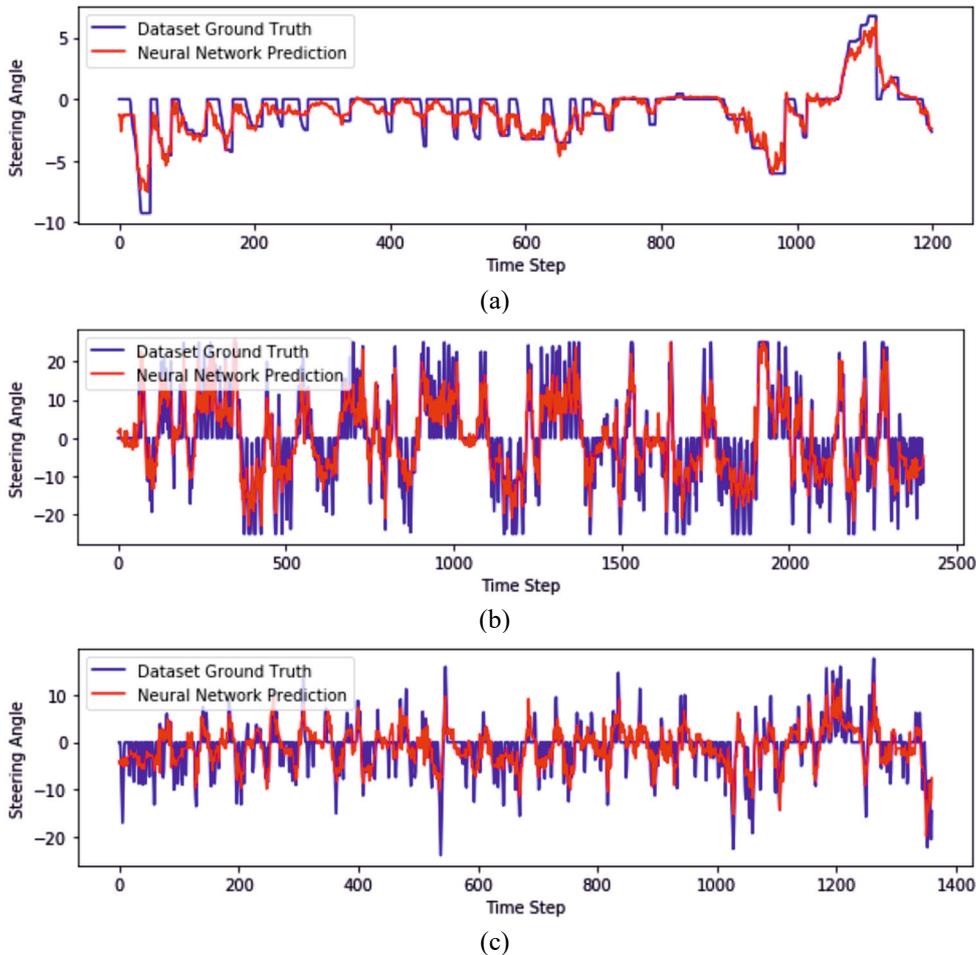

*Figure 13. Prediction analysis of (a) simplistic driving, (b) rigorous driving and (c) collision avoidance behavior models*

## 4.4. Simulated Field Results

Upon preliminary validation, as described in sections 4.2 and 4.3, the trained models were deployed onto the simulated ego vehicle so as to analyze the field results.

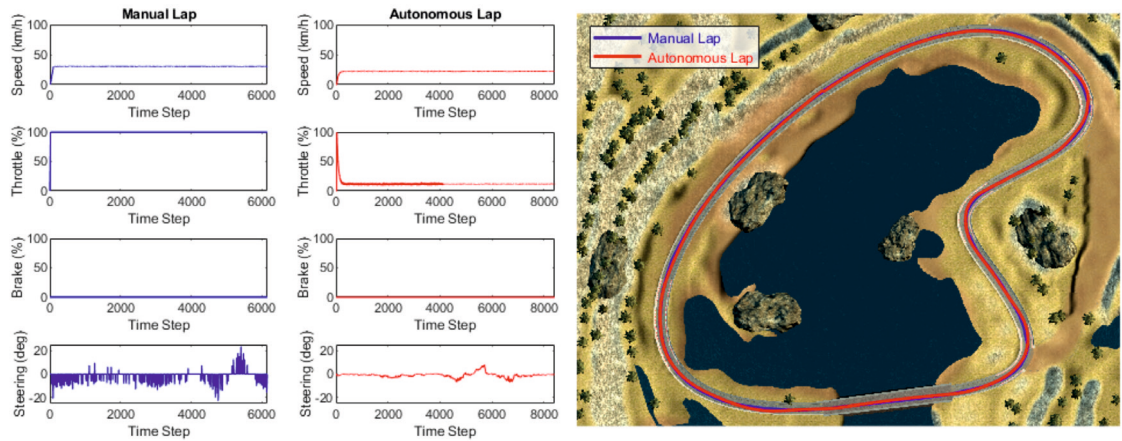

*Figure 14. Simulated field results of simplistic driving behavior model without environmental variations*

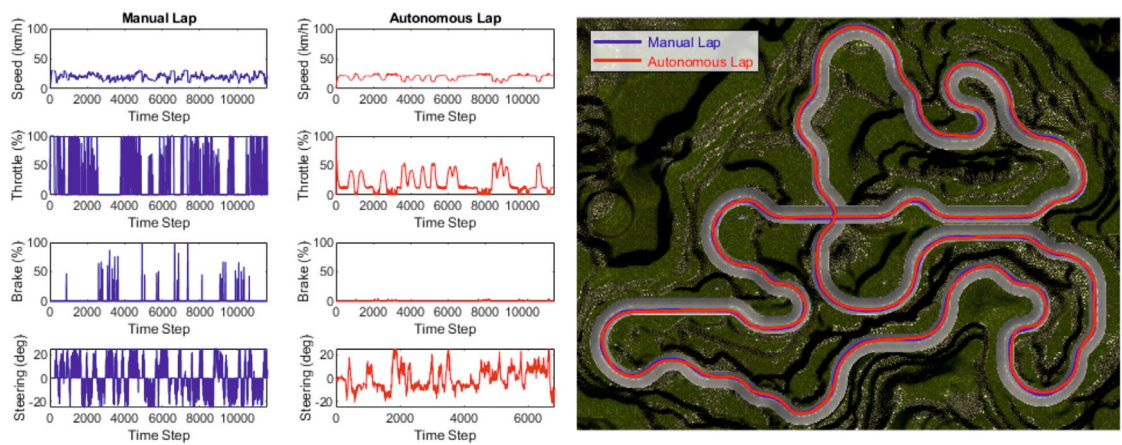

*Figure 15. Simulated field results of rigorous driving behavior model without environmental variations*

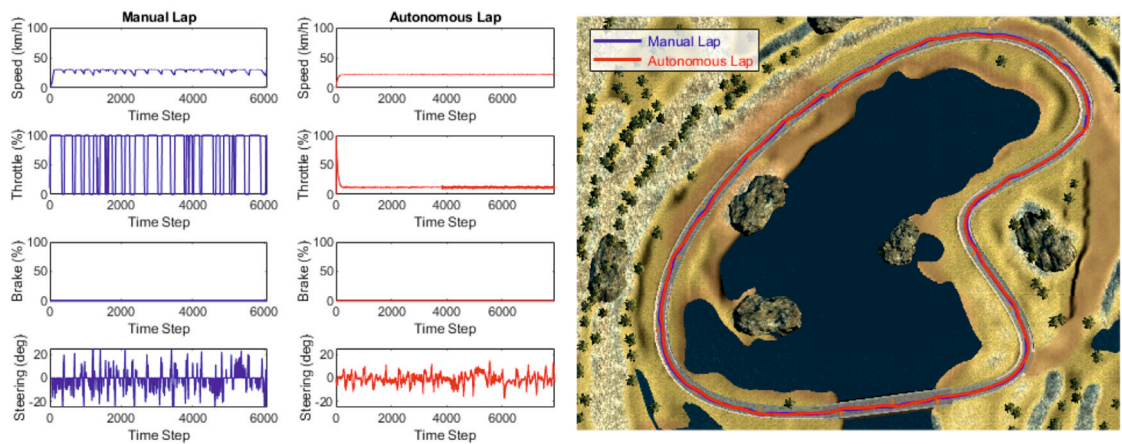

*Figure 16. Simulated field results of collision avoidance behavior model without environmental variations*

Initially, all the driving conditions were kept identical to those during data collection, with the only difference of a reduced speed limit of 25 km/h as compared to 30 km/h during manual driving. The trained network, along with the coupled control law (refer section 3.2.2) autonomously drove the vehicle across the respective driving scenarios. A comparative analysis was performed in order to analyze the degree of resemblance between manual and autonomous driving (Figures 14-16).

Next, the trained driving behavior models were subjected to robust testing as described in section 2.3. Table 9 summarizes the relative comparison of robustness metrics of the 3 driving behaviors. The collision avoidance behavior model proved to be the most robust owing to the fact that the course was pretty simple to maneuver and the model was trained particularly to avoid obstacles with aggressive maneuvers, which ultimately helped it recover from corner cases arising due to environmental variations. Next on the line was the rigorous driving behavior model, which despite being trained to handle sharp curves and bumps, occasionally wandered off the track solely due to the extremely challenging and arduous scene topography. Finally, the simplistic driving model performed most poorly in terms of exhibiting robustness to environmental variations, despite the simplicity of the driving scenario. The primary reason for this was that the model was trained for smooth driving and was not immune to corner cases, especially managing sharp turns by performing aggressive maneuvers.

*Table 9. Robustness metrics of various driving behaviors trained using our approach*

| Experiment | Robustness Measure[13] | | |
|---|---|---|---|
| | **Simplistic Driving** | **Rigorous Driving** | **Collision Avoidance** |
| No variation | $\eta \approx 100\%$ | $\eta \approx 100\%$ | $\eta \approx 100\%$ |
| Scene obstacle variation | … | … | $\eta\{20, 10, 0\} \approx 100\%$ |
| Scene light intensity variation | [-0.4 cd, 0.1 cd] | [-0.2 cd, 0.9 cd] | [-0.6 cd, 1.0 cd] |
| Scene light direction variation | [-14°, 1°] | [-2°, 4°] | [-18°, 28°] |
| Vehicle position variation | $\eta \approx 100\%$ | $\eta \approx 100\%$ | $\eta \approx 100\%$ |
| Vehicle orientation variation | [-35°, 15°] | [-25°, 25°] | [-55°, 40°] |
| Vehicle heading inversion | $\eta \approx 96.91\%$ | $\eta \approx 72.89\%$ | $\eta \approx 91.09\%$ |
| Vehicle speed limit variation | 40 km/h | 35 km/h | 50 km/h |

Finally, we compared our approach against NVIDIA's state-of-the-art implementation [9] in order to validate the performance of the proposed pipeline; results are summarized in Table 10. For this purpose, we adopted NVIDIA's PilotNet [10] architecture and trained it for cloning the simplistic driving behavior. We first trained a model using the pipeline described in [9] and then using our approach (all the training and deployment parameters were kept same as described earlier with the only exception of the neural network architecture and the corresponding resizing operation in the preprocessing pipeline so as to match the input size of the PilotNet: $200 \times 66$ px).

*Table 10. Comparative analysis of models trained using our approach against state-of-the-art*

| Parameter | Robustness Measure[13] | |
|---|---|---|
| | **NVIDIA's Approach [9]** | **Our Approach** |
| ***Computational Details*** | | |
| Training time | ~1.8 hr | ~1.4 hr |
| Deployment latency | 4-7 ms (mode = 4.99 ms) | 2-6 ms (mode = 3.99 ms) |
| ***Robustness Experiments*** | | |
| No variation | $\eta \approx 100\%$ | $\eta \approx 100\%$ |
| Scene light intensity variation | [-0.2 cd, 0.3 cd] | [-0.6 cd, 1.2 cd] |
| Scene light direction variation | [-3°, 7°] | [-30°, 30°] |
| Vehicle position variation | $\eta \approx 100\%$ | $\eta \approx 100\%$ |
| Vehicle orientation variation | [-20°, 10°] | [-45°, 10°] |
| Vehicle heading inversion | $\eta \approx 100\%$ | $\eta \approx 100\%$ |
| Vehicle speed limit variation | 50 km/h | 100 km/h |

A direct comparison between Table 8, 9 and 10 supports the claim of deeper neural networks possessing better generalization capability at the cost of increased training time and deployment latency. Taking a closer look, it can be observed that PilotNet trained using NVIDIA's approach [9] was only as robust as the relatively shallow network (refer section 3.1.6.1) trained using our approach, if not worse; not to mention the increased training time and deployment latency. On the other hand, our approach was able to train PilotNet much more robustly, within almost the same time as take by the shallower network. This validates our approach in terms of robust behavioral cloning for autonomous vehicles using end-to-end imitation learning. The slight increase in deployment latency can be attributed to the deeper network architecture with larger input size.

The video demonstrations for this work, pertaining to all the aforementioned experiments can be found at https://www.youtube.com/playlist?list=PLY45pkzWzH9-M6_ZBjynKyPlq5YsCzMCe. The simulator source code along with the training, analysis and deployment pipelines, all the datasets and trained neural network models for respective driving behaviors are released at https://github.com/Tinker-Twins/Robust_Behavioral_Cloning.

---

[13] The numerical values reported are an indicative of the robustness of trained models and are subject to minor variations owing to the probabilistic nature of the implementations.

# 5. CONCLUSION

This work presented a lightweight pipeline for training and deploying robust driving behavior models on autonomous vehicles using end-to-end imitation learning. The work also introduced a coupled control scheme so as to enhance the cooperative nature of lateral and longitudinal motion control commands. Additionally, a set of experiments and evaluation metrics for analyzing the efficiency and robustness of the proposed pipeline were formulated and presented as a part of this research. Three distinct driving behaviors were cloned using the proposed pipeline and exhaustive experimentation was carried out so as to test the bounds of the proposed system. Even a comparatively shallow neural network model was able to learn key driving behaviors from a sparsely labelled dataset and was tolerant to environmental variations during deployment of the said driving behaviors. Finally, the presented approach was validated by comparing it with NVIDIA's state-of-the-art implementation.

This work may be taken up to develop explicit hardware or sim2real implementations of end-to-end learning for autonomous driving. Additionally, the effect of collecting a diverse dataset from multiple human drivers and using substitute/multiple sensing modalities may be studied. Moreover, alternative approaches may be investigated to address the problem of generalization failure of end-to-end trained models in disparate scenarios. Furthermore, theoretical formulations for assessing reliability of autonomous systems trained using end-to-end learning may be researched exhaustively. Finally, this research may be pursued further in order to standardize the experiments and evaluation metrics for testing efficiency of an end-to-end learning pipeline and robustness of the trained models.